\documentclass[pdflatex,sn-mathphys-num,iicol,pagebackref=true]{sn-jnl}
\usepackage{multirow}%
\usepackage{amsmath,amssymb,amsfonts}%
\usepackage{amsthm}%
\usepackage{mathrsfs}%
\usepackage[title]{appendix}%
\usepackage{textcomp}%
\usepackage{manyfoot}%
\usepackage{booktabs}%
\usepackage{listings}%
\usepackage{url}
\usepackage{morefloats}
\usepackage{subcaption}
\usepackage{bchart}
\usepackage{pgfplots}
\usepackage{booktabs}
\usepackage{graphicx}
\usepackage[T1]{fontenc}
\usepackage{color,xcolor}
\usepackage{enumitem}
\usepackage{multirow}
\setcitestyle{numbers,square,citesep={,}} 
\theoremstyle{thmstyleone}%
\theoremstyle{thmstyletwo}%
\theoremstyle{thmstylethree}%

\newif\ifInDescription\InDescriptionfalse%
\newcommand\NestingFix{
  \ifInDescription%
    \leavevmode\vspace*{-\dimexpr2\topsep+\baselineskip\relax}%
  \fi%
}
\newlist{algorithm}{description}{1}
\setlist[algorithm]{style=nextline,                   before=\InDescriptiontrue,
                    after=\InDescriptionfalse,
                    font=\itshape\bfseries
}
\setlist{before=\NestingFix}
\usepackage{algorithm}
\usepackage{algpseudocode}
\begin{document}

\title[HW-MLVQA: Elucidating Handwritten Multilingual Document Understanding with a Comprehensive VQA Benchmark]{HW-MLVQA: Elucidating Multilingual Handwritten Document Understanding with a Comprehensive VQA Benchmark}

\author*{\fnm{Aniket} \sur{Pal}}\email{aniket.pal@research.iiit.ac.in}
\author{\fnm{Ajoy} \sur{Mondal}}\email{ajoy.mondal@iiit.ac.in}
\author{\fnm{Minesh} \sur{Mathew}}\email{minesh.mathew@gmail.com}
\author{\fnm{C. V.} \sur{Jawahar}}\email{jawahar@iiit.ac.in}
\affil{\orgdiv{CVIT}, \orgname{IIIT Hyderabad}}


\abstract{The proliferation of MultiLingual Visual Question Answering (MLVQA) benchmarks augments the capabilities of large language models (LLMs) and multi-modal LLMs, thereby enabling them to adeptly capture the intricate linguistic subtleties and visual complexities inherent across diverse languages. Despite its potential, the current MLVQA model struggles to fully utilize its capabilities when dealing with the extensive variety of handwritten documents. This article delineates \textbf{HW-MLVQA}, an avant-garde VQA benchmark meticulously crafted to mitigate the dearth of authentic Multilingual Handwritten document comprehension. \textbf{HW-MLVQA} encompasses an extensive collection of \textbf{\texttt{1,600 handwritten Pages}} complemented by \textbf{\texttt{2,400 question-answers}}. Furthermore, it provides a robust benchmark evaluation framework spanning \textbf{\texttt{three distinct modalities: text, image, and an integrated image \& text modality}}. To simulate authentic real-world contexts devoid of ground truth textual transcriptions, we facilitates a rigorous assessment of \textbf{\texttt{proprietary and open-source OCR models}}. The benchmark aspires to facilitate pivotal advancements in multilingual handwritten document interpretation, fostering innovation and scholarly inquiry within this specialized domain.}


\maketitle

\section{Introduction}

In present times, the domain of Visual Question Answering (VQA)~\cite{VQA,VQA_1} has witnessed remarkable advancements, accelerated by the thriving demand for methods capable of discerning and engaging with visual content through natural language. As an inherently interdisciplinary initiative, VQA amalgamates computer vision and natural language processing to elucidate and respond to inquiries about visual stimuli. Notwithstanding these developments, a predominant limitation persists wherein existing VQA frameworks predominantly cater to typed textual inputs and are confined to monolingual support, thereby engendering a noticeable difference in accessibility by primarily two obstacles: one, multilingualism, and two, interpreting contexts in complex handwritten format.

In order to mitigate the linguistic impediment inherent in vision-language tasks, Multilingual Visual Question Answering (ML-VQA)~\cite{MaXM,x_GQA,EVJVQA} was promulgated, thereby facilitating models to comprehend and answer questions articulated in a multitude of languages. For instance, Deepak \emph{et al.}\cite{mlvqa_code} pioneered an innovative dataset encompassing code-mixed Visual Question Answering (VQA) in both English and Hindi. Contemporary progressions have fostered the creation of datasets \cite{MaXM,x_GQA} characterized by refined annotation protocols, thereby augmenting the capability and applicability of VQA systems. Moreover, Nguyen \emph{et al.}~\cite{EVJVQA} have broadened the extent of VQA research to encompass linguistically underrepresented languages such as Vietnamese and Japanese.

In addressing the diverse handwriting complexity, the Handwritten VQA task \cite{HW-SQUaD} was conceptualized to facilitate the interpretation of handwritten documents, emphasizing complex and various handwriting styles. Two novel datasets, HW-SQuAD and Bentham-QA \cite{HW-SQUaD}, were introduced in conjunction with the task. More recently, to galvanize research interest and foster advancements within this field, an ICDAR competition \cite{Icdar_handwritten_challenge} was organized. Nevertheless, despite these initiatives, an evident gap persists—the absence of a comprehensive dataset that amalgamates the complexities of apprehending multilingual documents and diverse handwriting styles.

This study introduces \textbf{HW-MLVQA}, a comprehensive benchmark designed for handwritten multilingual visual question answering, to address the gap in multilingual document comprehension and the intricacy of handwriting. The benchmark enhances VQA systems' capabilities by enabling them to process and interpret handwritten questions across multiple languages.

\vspace{1mm}

Our key contributions to this work are:

\vspace{1mm}

\begin{itemize}

\item Introducing \textbf{HW-MLVQA}, a novel benchmark for handwritten multilingual document understanding.

\item Evaluating state-of-the-art models (LLM/VLM) across modalities, encompassing multilingual text models (LLaMA 3.1 ~\cite{llama}, M-BERT ~\cite{BERT}) and vision-language model (Qwen2VL \cite{Qwen2VL}), utilizing image-only and image-text inputs with OCR methods.

\item Assessing the visual grounding capabilities of Vision-Language Models to determine their proficiency in locating and identifying pertinent information within handwritten documents.

\end{itemize}

\section{Related works}\label{sec:related_work}

Significant progress has been made in multilingual VQA in recent years. For example, MLQA~\cite{mlqa} introduced a diverse, multi-way-aligned corpus spanning seven languages, while TyDi QA~\cite{tydiQA} expanded linguistic diversity by including 11 typologically distinct languages, tackling the challenges of building truly multilingual QA systems. More recently, the integration of visual and textual question answering has gained substantial attention. Datasets like TextVQA~\cite{TextVQA} and DocVQA~\cite{DocVQA} address the complexities of answering questions based on text embedded in images and documents. In 2023, Google introduced a foundational multilingual VQA dataset, and EVJVQA \cite{EVJVQA} was proposed to support resource-scarce languages like Japanese and Vietnamese. However, translation-based multilingual VQA datasets often face challenges like ``visual-textual misalignment,'' where only the textual aspects of question-answer pairs are considered, neglecting the visual text within images. To overcome this limitation, Multilingual Text-Centric VQA (MT-VQA) \cite{MTVQA} dataset was proposed to bridge the gap. 

Significant progress has also been made in handwritten document understanding. Notable datasets include IAM~\cite{IAM_dataset}, GNHK~\cite{lee2021gnhk}, and IIIT-HW-English-Word~\cite{mondal2024bridging} for English, RIMES~\cite{grosicki2011icdar} for French, and CASIA-HWDB~\cite{CASIA_HWDB} for Chinese, all of which have advanced OCR systems for these languages. In the handwritten VQA domain, datasets like HW-SQuAD~\cite{HW-SQUaD} and BenthamQA~\cite{HW-SQUaD} have shown potential. However, a comprehensive dataset specifically for handwritten VQA is still lacking.

Visual grounding in VQA has been advanced through datasets like Visual7W ~\cite{Visual7W}, GQA~\cite{GQA}, and VQA-HAT~\cite{vqahat}, which link questions and answers to image regions or objects. TextVQA focuses on text-based grounding, while CLEVR-Humans ~\cite{CLEVR} and ReferItGame ~\cite{ReferItGame} emphasize compositional reasoning and fine-grained phrase grounding. Real-world challenges are addressed in VizWiz ~\cite{VizWiz}, featuring images from visually impaired users, and OpenImages-VQA ~\cite{OpenImages}, combining large-scale tasks with object annotations. Despite these significant contributions, there remains a notable absence of visual grounding datasets explicitly addressing the unique challenges of handwritten multilingual documents.

\begin{figure*}[!ht]
\centerline{
\includegraphics[width=\textwidth]{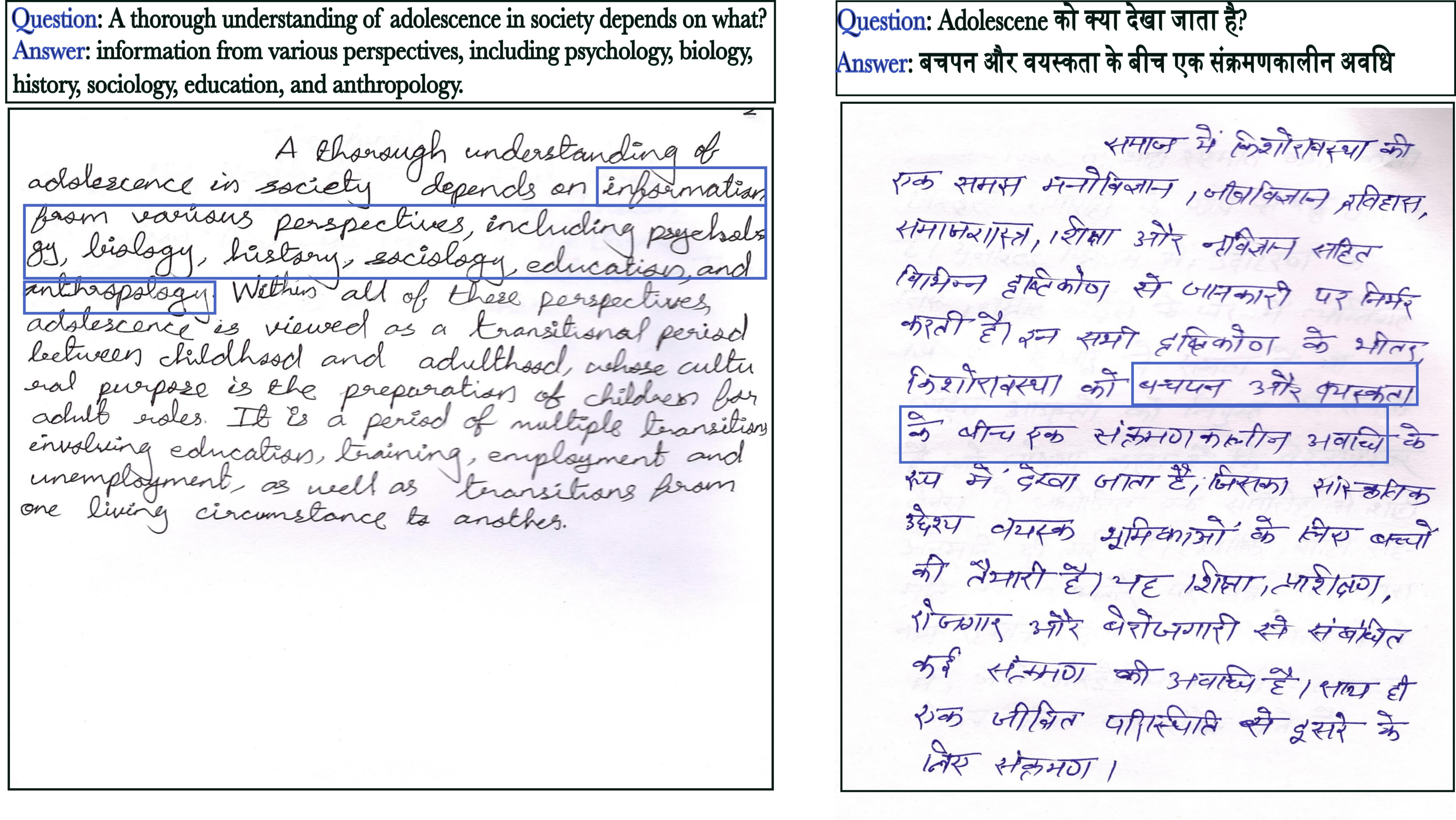}}
\caption{Examples of English (left) and Hindi (right) samples with QAs of the same context.\label{fig:example_hindi_english}}
\end{figure*}

Despite these advancements, a clear resource gap remains in handwritten multilingual VQA, particularly for languages with distinct scripts like Hindi and English. Currently, no benchmark address this need. To bridge this gap, we introduce the HW-MLVQA, proposing a unique benchmark for developing and evaluating systems that tackle the visual complexity of handwritten text and the linguistic challenges of multilingual visual question answering.

\begin{figure*}[!ht]
\centerline{
\includegraphics[width=\textwidth, height=0.3\textwidth]{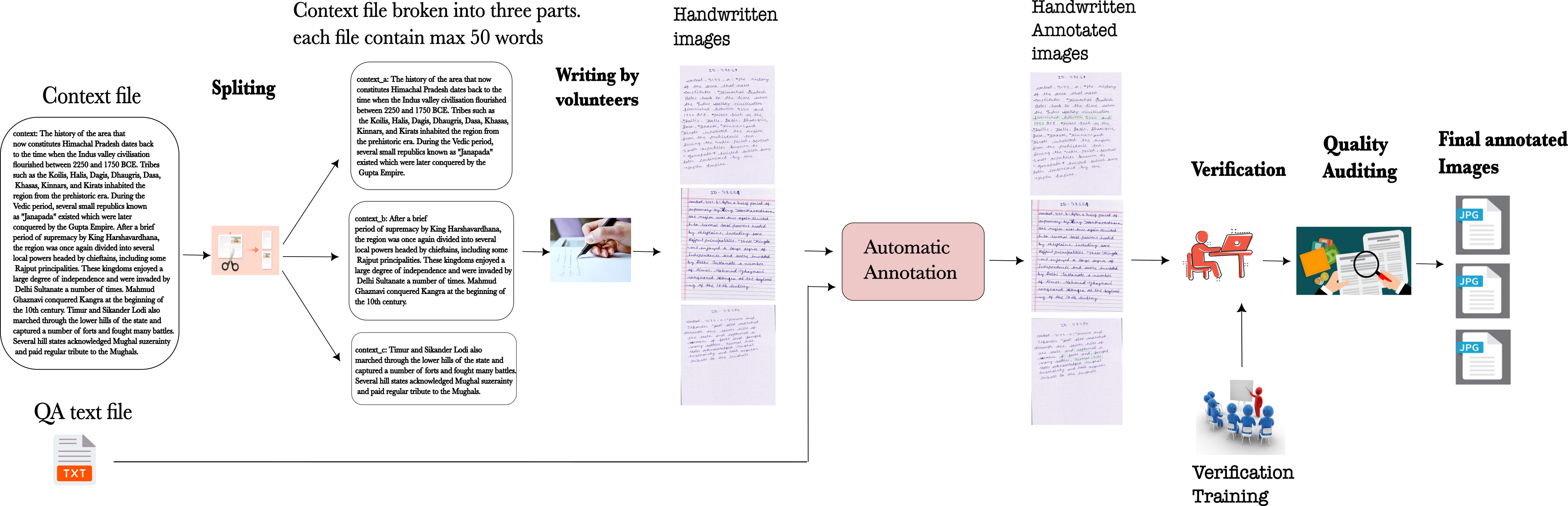}}
\caption{Shows the complete annotation pipeline involves several steps. First, the context files are divided into 50 word chunks and uploaded to a web-based tool. The handwritten copies are then scanned and processed through an automated annotation pipeline. Following this, the data undergoes verification and quality checks. Once these steps are completed, the final annotated images are collected.\label{fig:annotation_pipeline}}
\end{figure*}

\section{HW-MLVQA Benchmark}\label{sec:dataset} 

HW-MLVQA is a top-tier gold-standard evaluation benchmark, meticulously crafted to evaluate multimodal question answering capabilities across intricate handwritten multilingual documents. Its core objective revolves around \textbf{\texttt{Evidence-based, Grounded Visual Question Answering}}, ensuring comprehensive and precise assessments.

\subsection{Evidence-Based Grounded VQA} 

The Evidence-Based Visual Question Answering (EB-GVQA) task, encompassed within HW-MLVQA, evaluates a model's adeptness in retrieving, interpreting, and substantiating responses predicated upon handwritten multilingual evidence. In contrast to traditional multilingual VQA paradigms, which rely solely on text extracted through multilingual OCR, this task necessitates that models meticulously interpret handwritten visual semantics, accurately account for variances in writing, and adeptly manage the intricacies of multilingual handwriting variations. Moreover, it mandates that all responses be meticulously anchored to the evidence provided.

\subsubsection{Task-formulation} Upon receiving an input visual note, represented as a set of pages $I = \{I_v\}_{v=1}^{n}$ (where the number of pages $n \in \{1, 2, 3, 4\}$) and a natural language question, $Q$, the model is required to perform the following tasks:

\begin{enumerate}
    \item \textbf{Retrieve Relevant Evidence:} Accurately identify key segments within the pages of $I$ that are instrumental in formulating a response to $Q$.
    \item \textbf{Generate an Answer:} Craft a coherent natural language answer, $A$, derived from the retrieved evidence, $E$.
    \item \textbf{Provide Justification:} Clearly highlight the supporting evidence, $E$, within the input pages $\{I_v\}$, delineating the connection to the final answer.
\end{enumerate}

Formally, the model is defined as:
\begin{equation} \label{eq:model_def}
    A, E = f_{\text{EB-GVQA}}(\{I_v\}_{v=1}^{n}, Q)
\end{equation}
where the evidence set $E$ consists of:
\begin{equation} \label{eq:evidence_set}
    E = \{ (B_i, v_i) \}_{i=1}^{p}
\end{equation}
\begin{itemize}
    \item $B_i$ represents the bounding box of a relevant evidence region on page $I_{v_i}$.
    \item $v_i \in \{1, \dots, n\}$ is the index of the page containing the evidence.
    \item $p$ is the total number of evidence regions.
\end{itemize}

\subsubsection{Evaluation}

To provide a holistic assessment of model performance, we evaluate two key dimensions: answer accuracy, evidence retrieval quality.

\begin{itemize}
    \item[\textbf{1.}] \textbf{Answer Accuracy.} The correctness of the generated natural language answer is quantified using the \textbf{Average Normalized Levenshtein Similarity (ANLS)} \cite{DocVQA}. This metric is robust to minor OCR errors and is calculated over a dataset of $N$ samples as follows:
    \begin{equation} \label{eq:anls}
        \text{ANLS} = \frac{1}{N} \sum_{i=1}^{N} \left( 1 - \frac{L(A_{p_i}, A_{g_i})}{\max(|A_{p_i}|, |A_{g_i}|)} \right)
    \end{equation}
    where for the $i$-th sample, $A_{p_i}$ is the predicted answer, $A_{g_i}$ is the ground-truth answer, $L(\cdot, \cdot)$ denotes the Levenshtein distance (the minimum number of single-character edits required to change one string into the other), and $|\cdot|$ represents the string length. Both strings are typically lowercased and stripped of articles and punctuation before comparison.

    \item[\textbf{2.}] \textbf{Evidence Retrieval Quality.} The spatial accuracy of evidence localization is assessed via the \textbf{Intersection-over-Union (IoU)}. This metric calculates the overlap between the set of predicted evidence bounding boxes ($E$) and the ground-truth set ($E^*$).
    \begin{equation} \label{eq:iou}
        \text{IoU}(E, E^*) = \frac{|E \cap E^*|}{|E \cup E^*|}
    \end{equation}
\end{itemize}

As shown in Equation~\ref{eq:iou}, the IoU is a critical measure of model's grounding capability.

\subsection{Dataset creation and Annotation}

We began by selecting 4000 contexts from the SQuAD~\cite{rajpurkar2016squad} and MLQA~\cite{mlqa} datasets, focusing on those with the highest number of question-answer pairs. These contexts, available in both English and Hindi, served as the foundation for our bilingual dataset. To adapt these contexts for handwritten reproduction, we needed to break them down into smaller, manageable segments. Our analysis revealed that each handwritten page could comfortably accommodate about 50 words without overcrowding. As a result, we carefully divided each context into smaller sections, ensuring that no individual file exceeded the 50 word limit. This segmentation allowed us to maintain clarity and readability while adhering to handwritten text collection's practical constraints. By doing so, we ensured that the dataset would be suitable for both manual annotation and effective testing of handwritten multilingual VQA systems. Fig. \ref{fig:annotation_pipeline} illustrates the comprehensive end-to-end annotation pipeline.

\begin{figure*}[!ht]
\centerline{
\includegraphics[width=0.8\textwidth,height=0.4\textwidth]{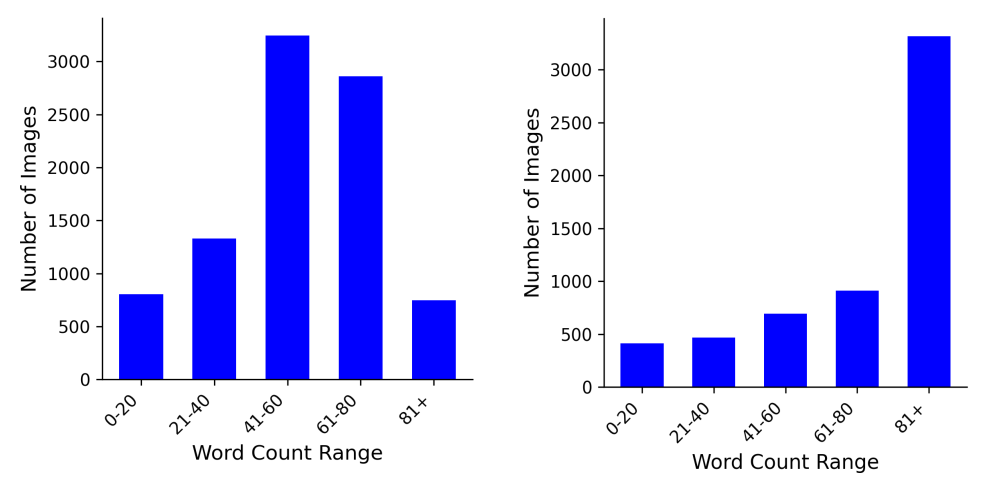}}
\caption{Shows word counts for English (left) and Hindi (right) images. The Hindi images contain more words per page than the English images.\label{fig:histogram}}
\end{figure*}

\subsubsection{Data Collection}

We developed a web-based platform to display segmented texts and recruited diverse volunteers to hand-transcribe them, ensuring dataset authenticity. Guidelines limited each handwritten page to 50 words for Hindi and English, using A4 paper with black or blue ink on ruled or unruled sheets. Random context assignments increased dataset variability.

Volunteers wrote at a natural pace to capture genuine handwriting. Quality control included spot checks and consistency assessments. Completed transcriptions were securely packaged, scanned, and digitised, preserving handwriting details. This comprehensive process provided a rich, authentic analysis and machine learning dataset.

\subsubsection{Annotation}

The annotation stage consists of two phases --- (i) automatic annotations are applied by aligning bounding boxes with the ground truth words using a heuristic algorithm, and (ii) a team is tasked with verifying the alignment of the bounding boxes with the words. The details of each procedure are outlined in the following section.

\vspace{0.2cm}

\noindent
\textbf{Automatic Annotation:} After data collection, we implemented an automatic annotation pipeline, which includes the following steps:

\begin{itemize}
\item \textbf{OCR Extraction:} We used two different APIs for Optical Character Recognition (OCR) to extract text from the scanned handwritten pages: a commercial API (Google OCR) and a freely available API (EasyOCR).

\item \textbf{Answer Matching:} After the OCR extracts the words, we segment them into lengths matching the expected answer. Using the SQuAD and MLQA datasets, we retrieve the responses from the QA files. The OCR extracted words are then compared with the query and the corresponding answers.

\item \textbf{Bounding Box Generation:} Matching OCR-extracted words with answer words, we generated bounding boxes to highlight answer locations. Each question-answer pair had a separate XML file with relevant coordinates. 
\end{itemize}

\begin{figure*}[!ht]
\centerline{
\includegraphics[width=\textwidth]{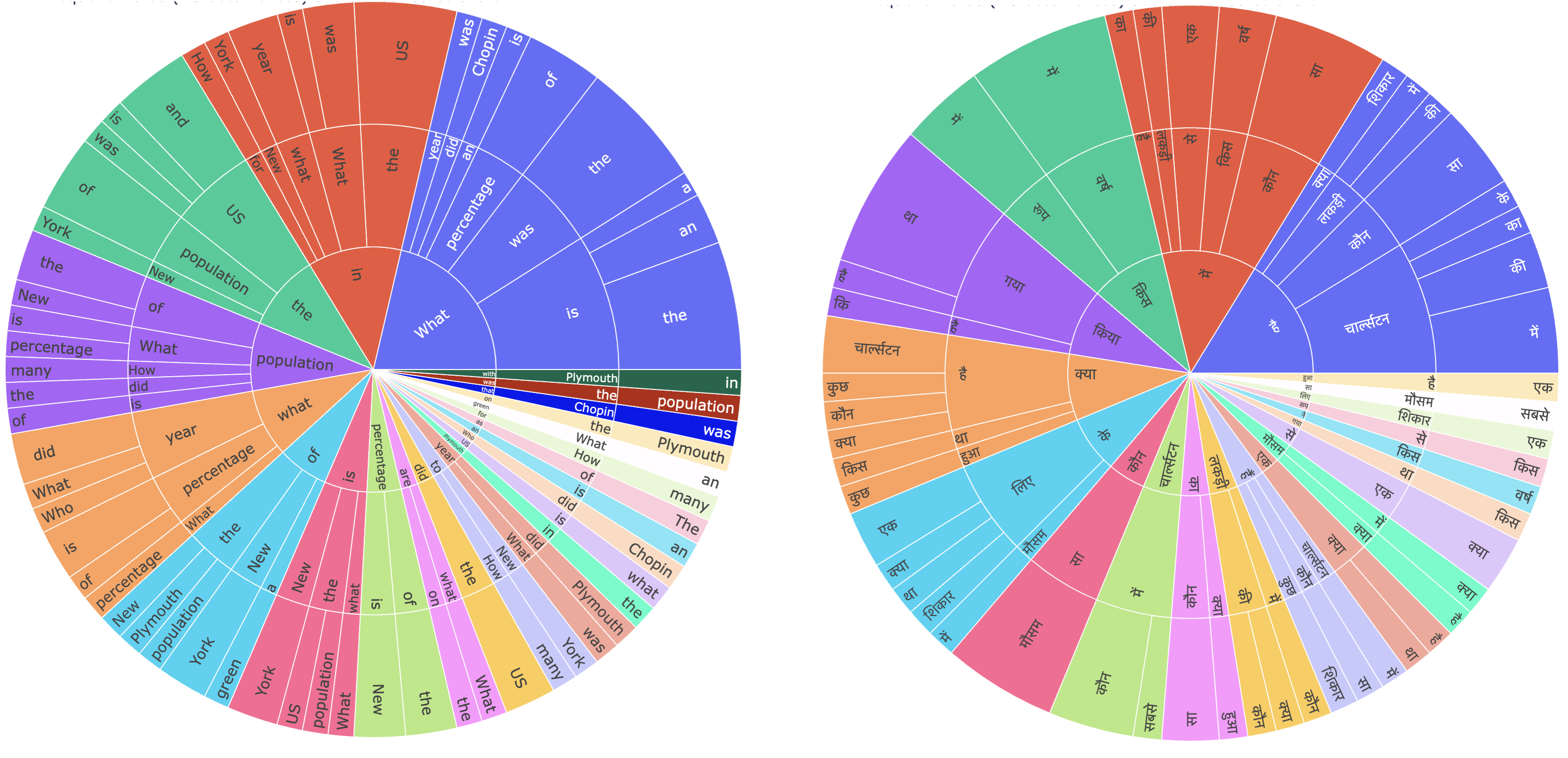}}
\caption{3-gram sunburst charts for questions in English (on the left) and Hindi (on the right), based on the first three words of each question.\label{fig:sunburst}}
\end{figure*}

\subsection{Annotation Verification}

We implemented a rigorous verification process to maintain high standards of quality and accuracy in our annotations. Using the LabelIMG tool, we saved XML files generated during the automatic annotation pipeline.

A trained team of five to six individuals was tasked with verification—two focused on Hindi annotations and the rest on English. They identified and corrected errors, ensuring accurate bounding boxes and consistent labeling across the dataset.

As a final quality assurance step, we conducted a thorough review to confirm the correctness and consistency of annotations, ensuring data reliability for machine learning applications.

\begin{figure*}[!ht]
\centerline{
\includegraphics[width=0.9\textwidth]{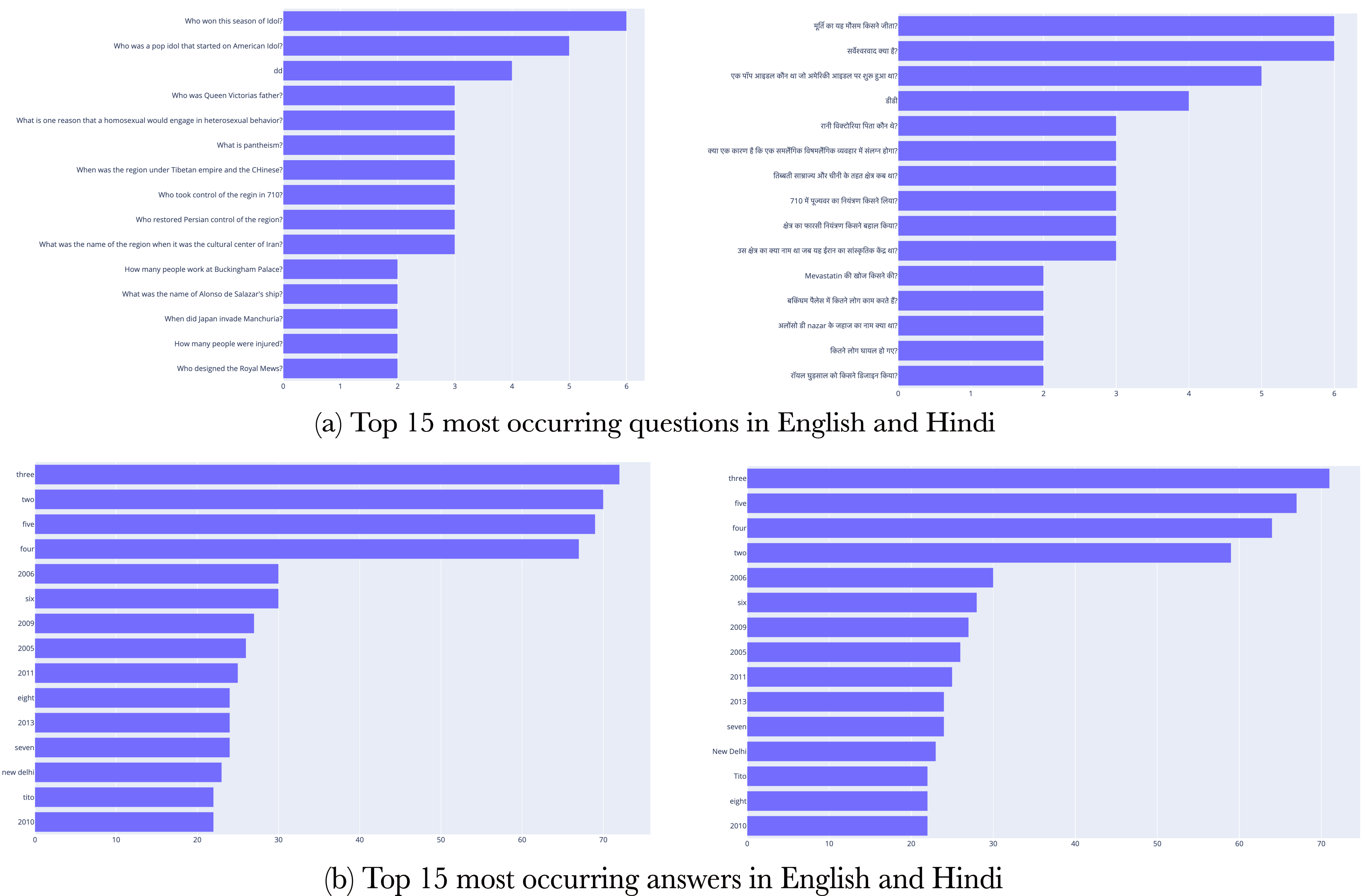}}
\caption{Shows statistics for questions and answers in both English and Hindi in the HW-MLQA dataset.\label{fig:top_15_qa}}
\end{figure*}

\subsection{Data Statistics and Analysis}\label{sec:Statistics}

 HW-MLVQA encompasses 2,400 questions and 1,520 Handwritten images. We also provide training and validation sets includes about 21,600 questions and 12,000 images. Table~\ref{tab:dataset-creation} shows the statistics of the ML-VQA dataset. Fig. ~\ref{fig:example_hindi_english} presents an sample of the benchmark showcasing English and Hindi content side by side along with their corresponding question-answer pairs.

\begin{table*}
\caption{Key statistics of the HW-MLVQA benchmark.}
\label{tab:dataset-creation}
\centering
\small
\begin{tabular}{@{}lrrrrrr@{}}
\toprule
\textbf{Language} & \textbf{Contexts} & \textbf{Avg. Words} & \textbf{Avg. Words/page} & \textbf{Pages}  & \textbf{Total number of}\\
 & & \textbf{per context} & & \textbf{(est.)} & {Questions} \\
\midrule
English & 400 & 116 & 60 & 834  & 2400\\
Hindi   & 400 & 136 & 80 & 686 & 2400\\
\bottomrule
\end{tabular}
\end{table*}

Figure \ref{fig:histogram} delineates the distribution of images across both English and Hindi languages concerning word count. Owing to the inherent linguistic intricacies of Hindi, more words are necessary to convey equivalent context compared to English. Fig.~\ref{fig:sunburst} showcases a sunburst plot illustrating the first three words of these questions. The plot reveals a variety of question types, including those starting with ``What,'' which often relate to inquiries where the answer is directly available in the text of the handwritten images. Fig.~\ref{fig:top_15_qa} (a) and (b) highlight the top 15 most frequently asked questions and their respective frequencies in both English and Hindi.

\begin{figure*}[!ht]
\centerline{
\includegraphics[width=\textwidth]{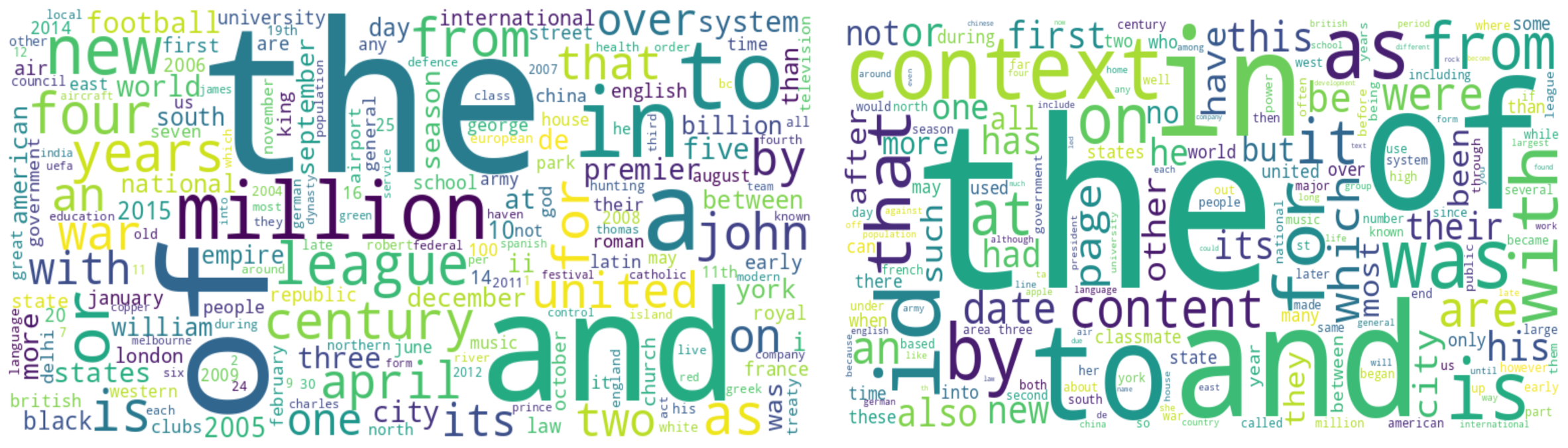}}
\caption{Present word clouds of English words in answers (left) and word clouds of English words in OCR tokens (right).\label{fig:word_cloud}}
\end{figure*}

\begin{figure*}[!ht]
\centerline{
\includegraphics[width=\textwidth]{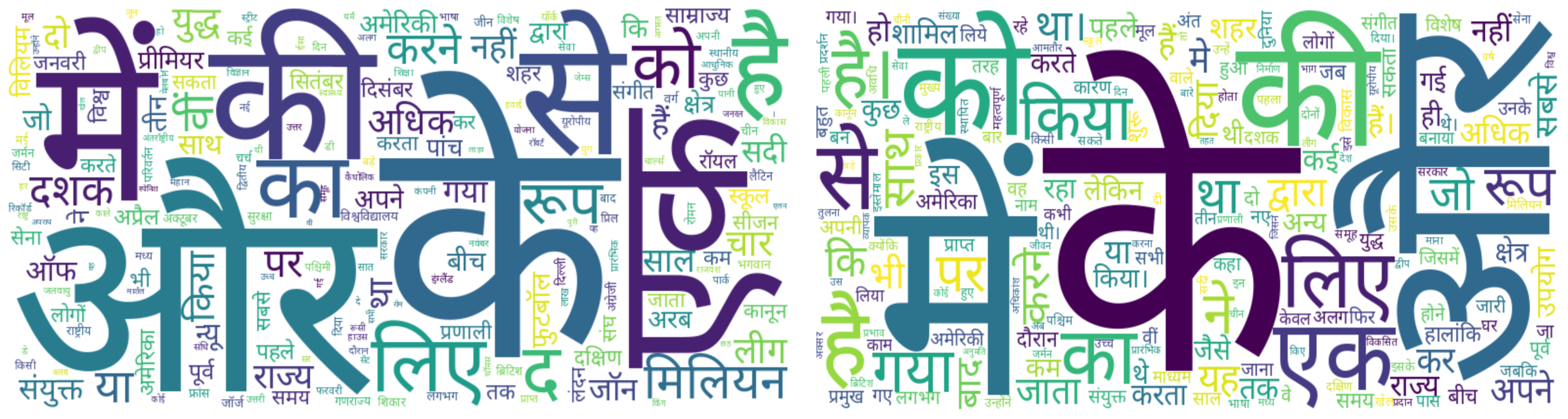}}
\caption{Shows word clouds of Hindi words in answers (left) and word clouds
of Hindi words in OCR tokens (right).\label{fig:word_hindi_cloud}}
\end{figure*}

The word clouds on the left side of Figs.~\ref{fig:word_cloud} and~\ref{fig:word_hindi_cloud} display the most common words found in the answers. These answers cover a wide range of information, such as names, amounts, months, and years. On the right side of the figures, the word clouds show the OCR tokens extracted from the images.

This extensive dataset provides a solid foundation for comprehensive model training and evaluation, enabling the development of robust systems capable of processing diverse linguistic patterns and visual inputs. The balanced data split ensures effective model tuning and unbiased performance evaluation. However, challenges may arise in maintaining annotation consistency and efficiently processing the large-scale multimodal data.

\section{Baseline Methods}

To establish a robust and comprehensive evaluation framework, we implement three baseline methods, each tailored to a specific data modality in our dataset ---(i) text (language), (ii) image (vision), and (iii) a combination of image and text (vision and language). These baselines serve multiple potential purposes: they act as performance benchmarks, provide modality-specific insights, contextualize the performance, evaluate robustness across varied input types, identify possible synergies between different data types, and facilitate error analysis. 

We evaluate model performance using linguistic, vision-based, and combined baselines. Linguistic baselines assess transcribed text from ground truth and OCR sources. Vision-based baselines focus on handwritten images. Combined baselines integrate both, offering insights into multimodal data handling. This approach reveals model robustness to OCR errors and the impact of transcription quality, enhancing our understanding of handwritten multilingual documents.

\subsection{Language Model}

We utilize LLaMA 3.1 8B ~\cite{llama}, a large language model with 8 billion parameters, optimized for multilingual tasks. Its 32-layer decoder, 32 attention heads, and extensive vocabulary (128,256 tokens) enable robust cross-lingual capabilities. Trained on diverse languages, LLaMA 3.1 excels in translation, sentiment analysis, multilingual content generation, and low-resource language support.

\subsection{Vision-Language Model} 

Our study exclusively used the Qwen2VL-7B model ~\cite{Qwen2VL} for evaluation. Developed by Alibaba Cloud's Qwen team, it features the Naive Dynamic Resolution mechanism for flexible, accurate image processing across varying resolutions. Qwen2VL excels in integrating visual and textual understanding, making it highly effective in tasks like image description and content comprehension.

\subsection{Evaluation Protocol}

We evaluate the performance of linguistic and vision-linguistic models in two different situations --- (i) when ground truth textual transcription is available and (ii) when ground truth textual transcription is not available.  

\subsubsection{Using Ground Truth Transcript}

Ground truth data is crucial for training and testing OCR systems, serving as a reliable reference for comparison. These datasets facilitate the objective evaluation of text recognition algorithms, mainly applied to diverse and complex documents. By leveraging ground truth data, the model performance can be accurately assessed without the influence of OCR errors.

\subsubsection{Using OCR Prediction}

To simulate real-world scenarios where text may be unavailable for handwritten documents, we evaluate both linguistic and vision-linguistic models using text extracted by two widely used OCR systems: (i) GoogleOCR – a commercial system known for its high accuracy and reliability, and (ii) EasyOCR – an open-source solution offering flexibility and adaptability across various applications. Table~\ref{tab:OCR_performance} presents the performance of GoogleOCR and EasyOCR on the HW-MLVQA dataset, focusing on word and character accuracy. This evaluation provides valuable insights into how OCR errors impact the model’s ability to understand and process text accurately.

\begin{table*}[!ht]
\caption{Shows the performance of commercial and non-commercial OCR on the ML-VQA dataset. Bold values indicate the best results.}
\label{tab:OCR_performance}
\centering
\small
\begin{tabular}{@{}llrrr@{}}
\toprule
\textbf{Language} & \textbf{OCR API} & \textbf{Word Accuracy}  & \textbf{Character Accuracy} \\
\midrule
English &GoogleOCR &\textbf{78.64} &\textbf{82.20}  \\
        &EasyOCR &5.56 &31.42  \\
\midrule
Hindi   &GoogleOCR  &\textbf{67.10} &\textbf{71.70}  \\
        &EasyOCR &8.29 &37.31  \\
\bottomrule
\end{tabular}
\end{table*}

\section{Experimental setup and Result analysis}

\subsection{ Experimental Setup}

The datasets were meticulously curated, encompassing OCR outputs, ground truth annotations, and question-answer pairs formatted in the SQuAD JSON structure for both LLM and VLM. In the context of VLMs, handwritten pages were incorporated as image-based chat prompts for each question-and-answer set, supplementing the OCR outputs. All experimental procedures were executed on Nvidia V100 GPUs.

\subsection{Result Analysis}

\subsubsection{Evaluation Metrics}

We use three evaluation metrics --- \textbf{Exact Match (EM)}~\cite{DocVQA}, \textbf{F1 score}, and \textbf{Average Normalized Levenshtein Similarity (ANLS)}~\cite{DocVQA}. Exact Match calculates the percentage of questions where the predicted answer exactly matches the target answer, word for word. The F1 Score, a harmonic mean of precision and recall, assesses the balance between correctly predicted answers and the total number of relevant answers, making it especially useful for imbalanced datasets. ANLS (Average Normalized Levenshtein Similarity) is a similarity-based metric that allows for minor mismatches, such as those caused by OCR errors and uses Levenshtein distance to measure how closely the predicted answer aligns with the target.

\begin{table*}[!ht]
\caption{Shows the performances of linguistic models under zero-shot settings on English and Hindi test sets in three different situations. The bold values indicate the best results.}
\label{tab:result_llama}
\centering
\small
\begin{tabular}{@{}llrrrr@{}}
\toprule
\textbf{Model} & \textbf{Transcription} & \textbf{F1 score} & \textbf{EM} & \textbf{ANLS} \\
\midrule
\multirow{3}{*}{LLaMA-3.1 8B (Test-En)}  &EasyOCR & 17 & 11 & 35.00\\
 & GoogleOCR &50 &33 &63.33\\
 & Ground Truth &\textbf{65} &\textbf{48} &\textbf{75.00}\\
\midrule
\multirow{3}{*}{LLaMA-3.1 8B (Test-Hi)} &Easy OCR &2 &1.1 &12 \\
 &GoogleOCR &19 &11.0 &33\\
 &Ground truth &\textbf{33} &\textbf{20.0} &\textbf{41}\\
\bottomrule
\end{tabular}
\end{table*}

\begin{table*}[!ht]
\caption{Shows the performance of large vision-language models under zero-shot settings on English and Hindi test sets in seven different situations. The bold values indicate the best results.}
\label{tab:result_qwen}
\centering
\small
\begin{tabular}{@{}lrrlrrrr@{}}
\toprule
\textbf{Model} & \textbf{Image} & \textbf{Text} & \textbf{Text type} & \textbf{F1 score} & \textbf{EM} & \textbf{ANLS} \\
\midrule
\multirow{7}{*}{Qwen2VL-7B (Test-En)} & - & \checkmark &EasyOCR &5.37 &3.18 & 5.27\\
 & - & \checkmark & GoogleOCR &61.94 &46.70 &61.02\\
 & - & \checkmark &Ground Truth &80.39 &67.21 &75.89 \\
 & \checkmark & - & - &71.32 &57.51 &69.11\\
 & \checkmark & \checkmark &EasyOCR &68.76 &54.84 &66.00\\
 & \checkmark & \checkmark &GoogleOCR &71.10 &56.53 &69.21\\
 & \checkmark & \checkmark &Ground Truth &\textbf{81.21} &\textbf{68.00} &\textbf{76.89}\\
\midrule
\multirow{7}{*}{Qwen2VL-7B (Test-Hi)} & - & \checkmark &EasyOCR & 6.64 & 3.94 & 8.09\\
 & - & \checkmark &GoogleOCR & 19.42 & 13.55 & 19.46\\
 & - & \checkmark &Ground Truth & 34.21 & 24.59 & 31.30\\
 & \checkmark & - & - & 29.22 & 22.70 & 28.97\\
 & \checkmark & \checkmark &EasyOCR & 27.76 & 21.51 & 27.55\\
 & \checkmark & \checkmark &GoogleOCR & 32.76 & 25.09 & 31.99\\
 & \checkmark & \checkmark &Ground Truth &\textbf{41.66} &\textbf{31.94} &\textbf{38.45}\\
\bottomrule
\end{tabular}
\end{table*}

We analyze the performance of the model across three distinct scenarios: (i) when provided with linguistic information (text) as input, (ii) when provided with visual information (image) as input, and (iii) when both linguistic and visual information (text and image) are combined as input. Furthermore, we examine the impact of utilizing different modalities on the performance of vision-language model. The model's capability to localize answers within handwritten pages has also been evaluated.

\subsubsection{Results of Linguistic-based Model} 

This section presents the evaluation results of the text-based model, LLaMA 3.1 (8B), on ground truth transcriptions and text extracted using two OCR systems from handwritten documents. The results are summarized in Table~\ref{tab:result_llama}, with the first row outlining the model's performance on English test set using ground truth, EasyOCR, and GoogleOCR outputs. 

\begin{figure*}[!htbp]
\centerline{
\includegraphics[width=0.8\textwidth]{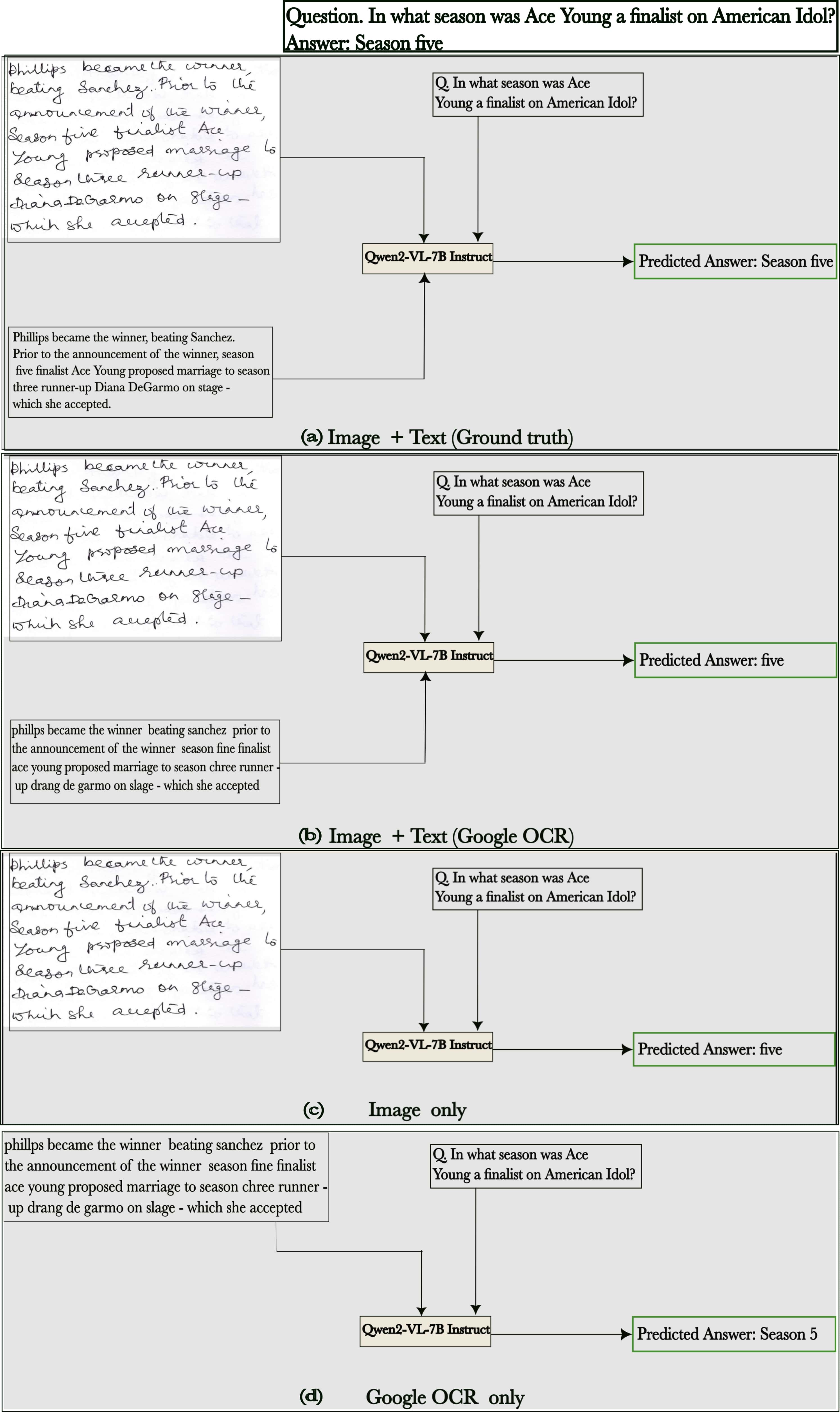}}
\caption{Presents the impact of different modalities on Qwen2VL-7B Instruct (English): comparison of combined image and text input (first two) versus image or text only input (latter two).}
\label{fig:qwen2vl_english}
\end{figure*}

\begin{figure*}[!htbp]
\centerline{
\hspace*{-2cm}\includegraphics[width=0.83\textwidth, height = 20cm]{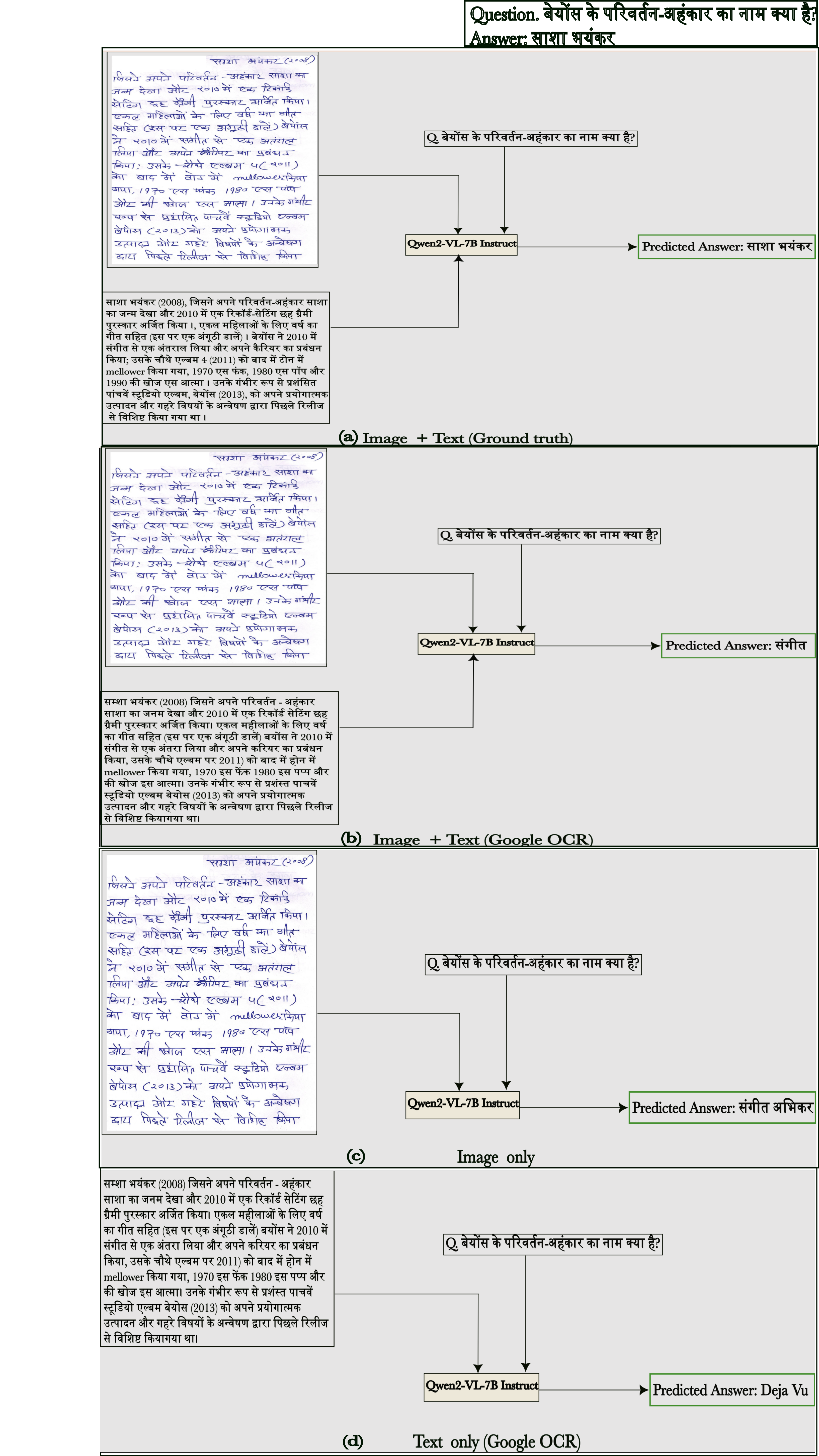}}
\caption{Presents the impact of different modalities on Qwen2VL-7B Instruct (Hindi): comparison of combined image and text input (first two) versus image or text only input (latter two).}
\label{fig:qwen2vl_hindi}
\end{figure*}

In a zero-shot setting, using EasyOCR, LLaMA 3.1 achieves an Exact Match (EM) score of 11\%, an F1 score of 17\%, and an ANLS score of 35\%. Similarly, with GoogleOCR, the model attains an EM score of 33\%. In contrast, when evaluated on ground truth transcription, the model achieves an EM score of 48\%, representing the upper performance bound for the English portion of our dataset using the LLaMA 3.1 model.

The state-of-the-art large vision-language model, Qwen2VL, was also evaluated using text-only inputs. The model achieved an EM score of 67.21\% on English ground truth data, while on Hindi ground truth data, it achieved 22.70\%. The model's performance dropped significantly when provided with text extracted using EasyOCR, achieving only approximately 3\% EM. In contrast, when tested with text extracted using GoogleOCR, the model attained EM scores of 46.70\% for English and 13.55\%  for Hindi.

The differences in performance between the two OCR datasets primarily reflect the quality of the OCR outputs, with EasyOCR performing notably worse than Google OCR. Consequently, the model achieves at least 20\% lower EM scores on English test data when relying on OCR outputs compared to the ground truth.

In the case of Hindi test data, the model's performance is significantly affected by the inherent complexity of the Hindi language. With the same number of examples as the English test set, the model struggles to capture the structural nuances of Hindi, performing less effectively than it does with English. It indicates that the model, optimized primarily for English, does not generalize well to Hindi, emphasizing the need for further adaptation or fine-tuning to handle linguistically diverse datasets effectively. 

\begin{table*}[!ht]
\caption{Show the result of grounding the answers in English and Hindi test sets.}
\label{tab:result_grounding}
\centering
\small
\begin{tabular}{@{}lrrrrr@{}}
\toprule
\textbf{Model} & \textbf{Max IoU} & \textbf{Avg (or Mean) IoU} & \textbf{Min IoU}  & \textbf{Var of IoU}\\
\midrule
\multirow{1}{*}{Qwen2VL-7B instruct (Test-En)}  & 0.6631 & 0.0166 & 0.0000 & 0.00197\\
\midrule
\multirow{1}{*}{Qwen2VL-7B instruct (Test-Hi)} & 0.2765 & 0.0126 & 0.0000 & 0.001055\\
\bottomrule
\end{tabular}
\end{table*}

\subsubsection{Results of Vision based Models}

Table~\ref{tab:result_qwen} depicts the results (The image only check mark). For the English handwritten image dataset, the model achieves an EM score of 45.39\%, an F1 score of 60.13\%, and an ANLS of 66\%. In contrast, the model obtains an EM score of 22.70\% for the Hindi handwritten image dataset, highlighting the greater linguistic complexity of Hindi compared to English. The image-only input evaluates the model's vision-based performance without supplementary information, such as OCR-extracted text. These experiments demonstrate that state-of-the-art vision-language models like Qwen2VL struggle to effectively capture linguistic structures solely from handwritten image-based samples in a zero-shot setting. It indicates that significant opportunities for improvement remain for large vision-language models like Qwen2VL.

\begin{figure*}[!ht]
\centerline{
\includegraphics[width=0.7\textwidth,height=0.4\textwidth]{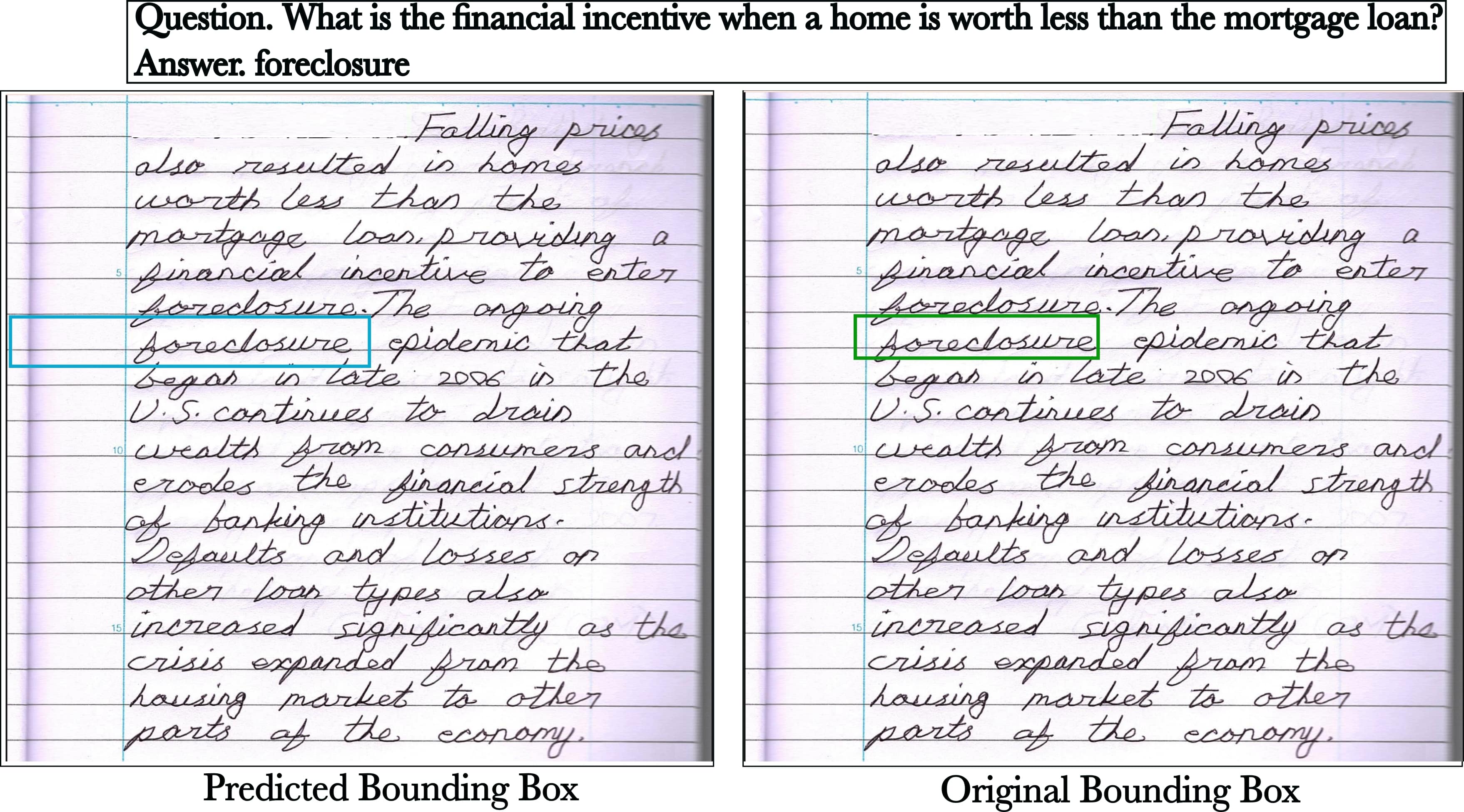}}
\caption{Shows the predicted and ground truth bounding box of answer on English test data with an IoU of 0.66.}
\label{fig:qwen2vl_bounding_box_english}
\end{figure*}

\begin{figure*}[!ht]
\centerline{
\includegraphics[width=0.7\textwidth,height=0.4\textwidth]{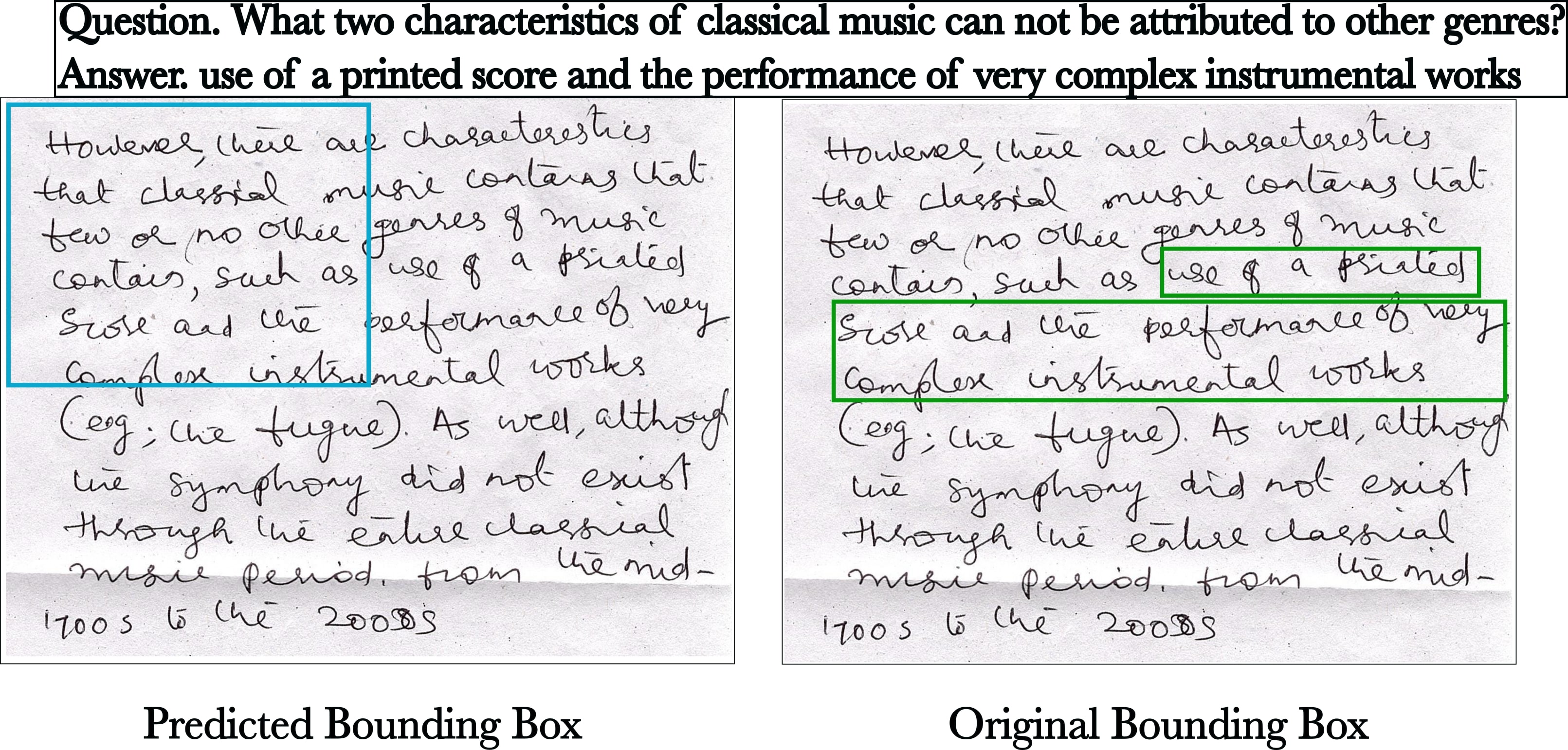}}
\caption{Shows the predicted and ground truth bounding box of answer on English test data with an IoU of 0.0719.}
\label{fig:qwen2vl_bounding_box_low_iou}
\end{figure*}

\begin{figure*}[!ht]
\centerline{
\includegraphics[width=0.7\textwidth,height=0.4\textwidth]{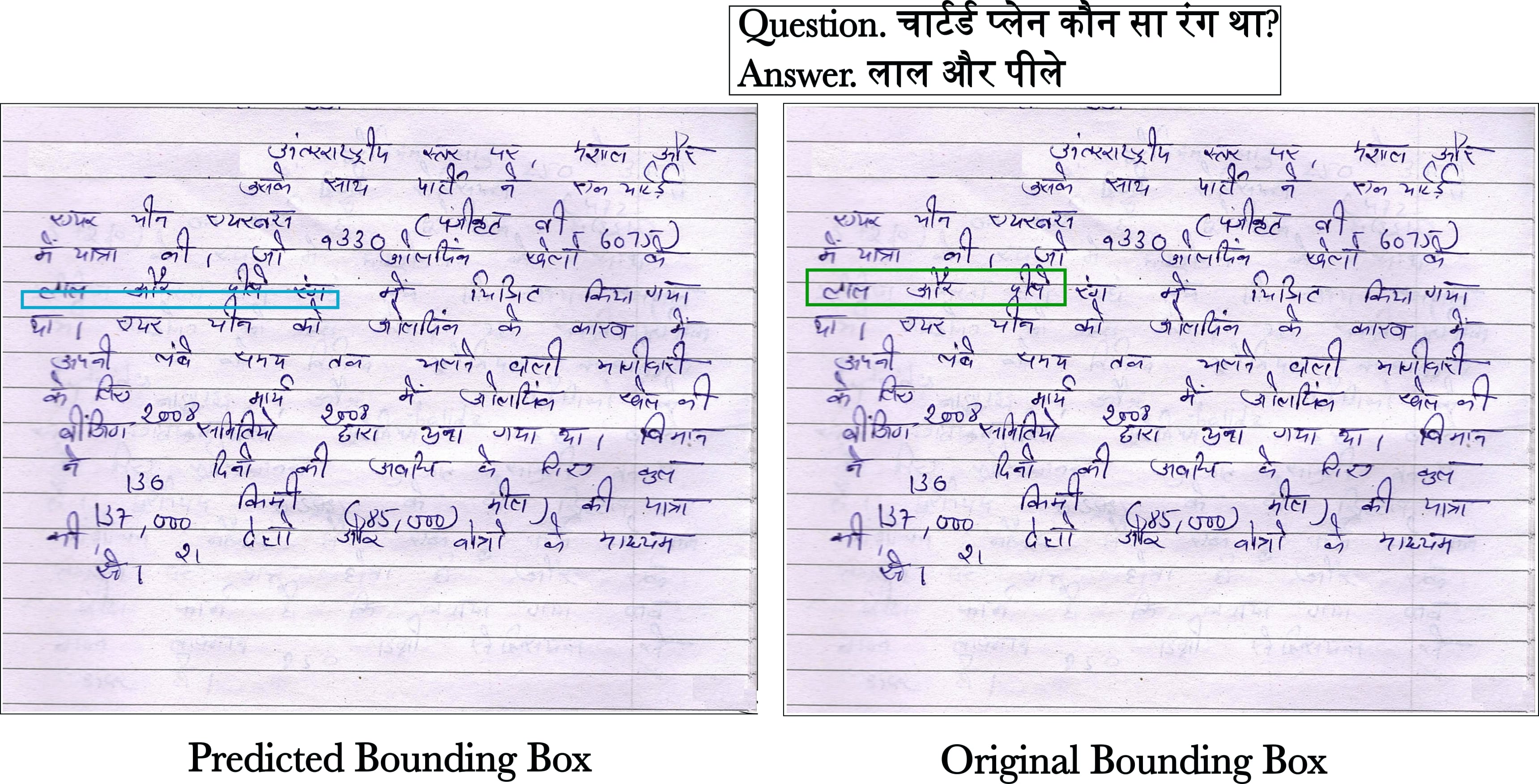}}
\caption{Shows the predicted and ground truth bounding box of answer on Hindi test data with an IoU of 0.27.}
\label{fig:qwen2vl_bounding_box_hindi}
\end{figure*}

\begin{figure*}[!ht]
\centerline{
\includegraphics[width=0.7\textwidth,height=0.4\textwidth]{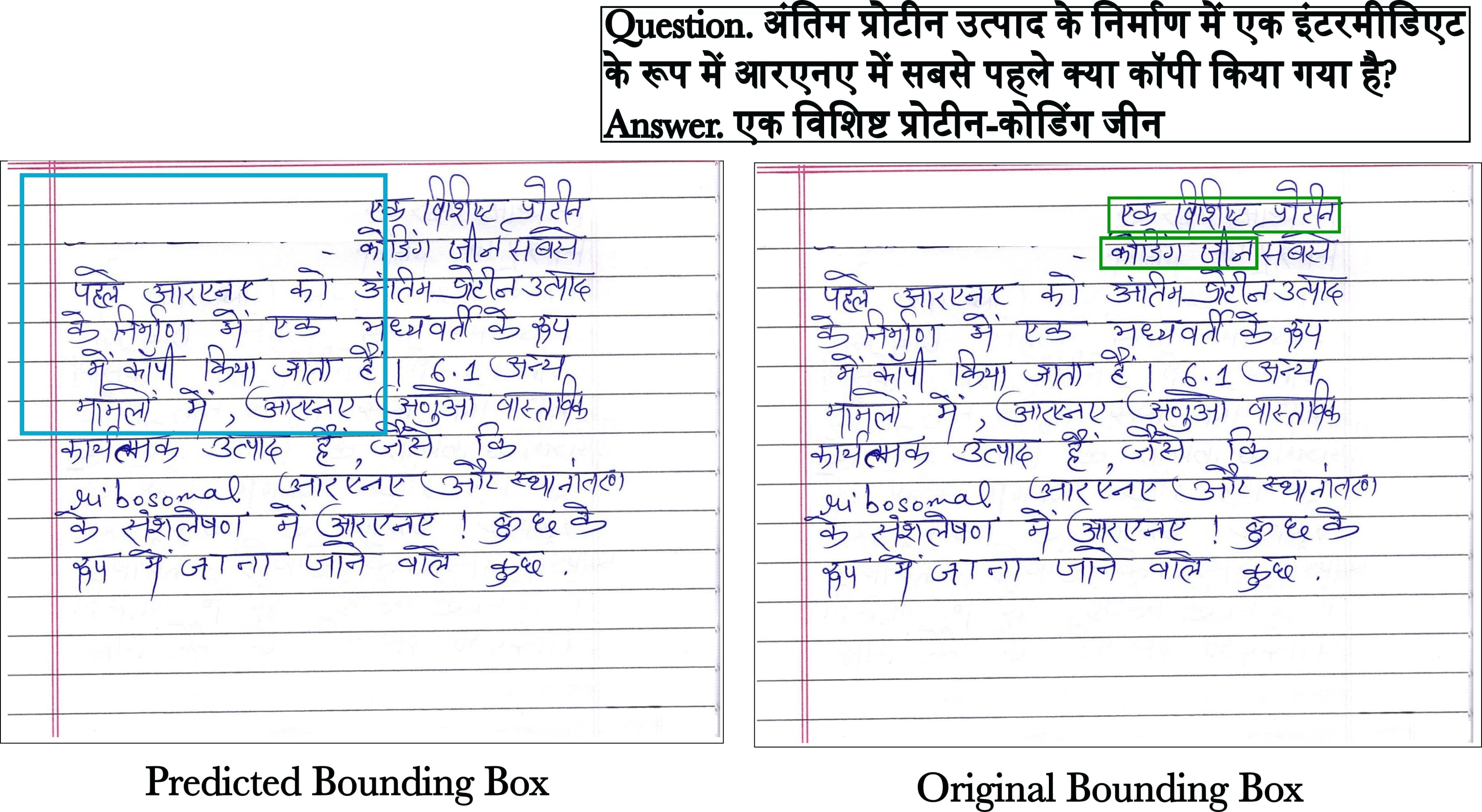}}
\caption{Shows the predicted and ground truth bounding box of answer on Hindi test data with an IoU of 0.0110.}
\label{fig:qwen2vl_bounding_box_hindi_low_iou}
\end{figure*}

\subsubsection{Results of Vision-Linguistic Models}

Table~\ref{tab:result_qwen} summarizes the findings, showcasing performance metrics across image and text inputs.
For English, the model is evaluated using prompts that include an image and its corresponding text derived from EasyOCR, GoogleOCR, or ground truth data. The model achieves an EM of 68.00\%, an F1 score of 81.21\%, and an ANLS of 76.89\% when ground truth text is used. When using EasyOCR, the EM score drops to 54.84\%, while GoogleOCR results in an EM score of 56.53\%.
For the Hindi dataset, the model achieves an EM score of 31.94\%, an F1 score of 41.66\%, and an ANLS score of 38.45\%. Like the English dataset, the model is evaluated with text derived from EasyOCR and GoogleOCR. In this case, the EM scores are 21.51\% and 25.09\%, respectively.
These experimental results highlight a significant degradation in the model performance when OCR errors are present, often leading to hallucinations in the model output. The highest performance, particularly regarding the exact match, is observed when ground truth data is provided, underscoring the importance of accurate text inputs for optimal results.

\subsubsection{Impact of Modalities on Vision-Language Model Performance}

This section examines the impact of image-text combinations and standalone text input on the Qwen2VL model's performance. Fig.~\ref{fig:qwen2vl_english} and Fig~\ref{fig:qwen2vl_hindi} present the results for English and Hindi, respectively. In Fig.~\ref{fig:qwen2vl_english}, the first two rows illustrate the outcomes when image and text modalities are combined. In comparison, the last two rows show the results when the image or text modality is provided as independent input to the model.
When provided with a handwritten image alongside the ground truth text, the Qwen2VL 7B model predicts the ground truth answer, ``Season five.'' However, when the image is paired with text generated by Google OCR, the model predicts ``five'' because the OCR misinterprets the text as ``Season fine'' instead of ``Season five.'' When only the image is provided, the model predicts ``five,'' whereas when only the Google OCR text is supplied, the model predicts ``Season 5.''

For Hindi, the evaluation was conducted using four distinct modalities (Fig.~\ref{fig:qwen2vl_hindi}). Due to the complexity of Hindi compared to English, the model struggles to accurately capture the linguistic structure when provided with inputs consisting solely of text or images. When images paired with Google OCR text are used as input, the model fails to predict the correct answer, producing outputs that are neither close to the ground truth nor any substring. However, the model successfully predicts the correct answer when the image and ground truth text are provided.

\subsubsection{Evaluation of Handwriting Localization in Vision-Language Model}

This section evaluates the performance of the Vision-Language Model, Qwen2-VL, in localizing answers within images from the test set. The model achieves a mean Intersection over Union (IoU) of 0.0166 for English handwritten data, with a variance of 0.00197. Fig.~\ref{fig:qwen2vl_bounding_box_english} and Fig.~\ref{fig:qwen2vl_bounding_box_hindi} illustrate the highest IoU scores achieved for English and Hindi, respectively. Fig.~\ref{fig:qwen2vl_bounding_box_hindi} and Fig.~\ref{fig:qwen2vl_bounding_box_hindi_low_iou} provide examples of instances with significantly lower IoU scores in both languages. The results indicate that the IoU is substantially lower for most samples, with the mean IoU for Hindi being 0.0126 lower than that for English, likely due to the greater complexity of Hindi handwriting than English. Table~\ref{tab:result_grounding} presents a comprehensive summary of the results. This analysis highlights the significant limitations of document-specialized VLMs, such as Qwen2-VL, in handwriting localization tasks and emphasizes further research to develop more effective models for addressing this challenge.

\section{Conclusion}

In this study, we introduce Visual Question Answering and visual grounding benchmark for handwritten multi-lingual documents and provide baseline performance using state-of-the-art Large Language Models (LLMs) and Large Vision-Language Models (LVLMs). Furthermore, we provide result analysis by simulating real-world scenarios where ground truth annotations are unavailable, with the ground truth serving as an upper bound for performance on the dataset. This research aims to pave the way for new advancements in handwritten multilingual VQA in future.

\bibliography{reference}

\end{document}